\documentclass[journal]{IEEEtran}
\IEEEoverridecommandlockouts
\pdfoutput=1
\usepackage{cite}
\usepackage{units}
\usepackage{lipsum}
\usepackage[caption=false]{subfig}
\usepackage{graphicx}	
\usepackage{pstricks}
\usepackage{float}
\usepackage{eurosym}
\usepackage{multicol}
\usepackage{booktabs}
\usepackage{units}
\usepackage{amsmath,amssymb}
\usepackage{pifont}
\usepackage{gensymb}

\graphicspath{{Figure/}}
\begin{document}

\title{Design of a prototypical platform for autonomous and connected vehicles}


\author{\IEEEauthorblockN{S. Arrigoni$^{1}$, S. Mentasti$^{2}$, F. Cheli$^{1}$, M. Matteucci$^{2}$, F. Braghin$^{1}$}

%\thanks{The Italian Ministry of Education, University and Research is acknowledged for the support provided through the Project "Department of Excellence LIS4.0 - Lightweight and Smart Structures for Industry 4.0”.}

\thanks{$^{1}$ S. Arrigoni, F. Cheli, F.Braghin are with the Department of Mechanical Engineering of Politecnico di Milano, via La Masa 1, Milan, Italy, {\tt\small name.surname@polimi.it}}%
\thanks{$^{2}$ S. Mentasti, M. Matteucci are with the Department of Electronics Information and Bioengineering of Politecnico di MIlano, p.zza Leonardo da Vinci 32, Milan, Italy, {\tt\small name.surname@polimi.it}}%

%\thanks{The results presented in this paper were obtained within the research project BUS-no-STOP (ID 30220968 - CUP code E47I11000720004) co-funded by Regione Lombardia.}
}


\maketitle

\begin{abstract}
Self-driving technology is expected to revolutionize different sectors and is seen as the natural evolution of road vehicles. In the last years, real-world validation of designed and virtually tested solutions is growing in importance since simulated environments will never fully replicate all the aspects that can affect results in the real world.
To this end, this paper presents our prototype platform for experimental research on connected and autonomous driving projects. In detail, the paper presents the overall architecture of the vehicle focusing both on mechanical aspects related to remote actuation and sensors set-up and software aspects by means of a comprehensive description of the main algorithms required for autonomous driving as ego-localization, environment perception, motion planning, and actuation. 
Finally, experimental tests conducted in an urban-like environment are reported to validate and assess the performances of the overall system.
\end{abstract}

\IEEEpeerreviewmaketitle

\section{introduction}

Autonomous vehicles (AV) are one of the most disruptive and impactful technologies of this century. Self-driving technology is expected to revolutionize different sectors and is seen as the natural evolution of road vehicles. Shortly AV technologies will become more and more accessible and familiar in everyday life and enable cheaper, safer, and more accessible means of transport. Indeed, cars are one of the most dangerous ways of transportation. According to the European Commission report on road safety~\cite{safetyEUroad} 25100 people died in 2018 and many more remained severely injured for road traffic fatalities just in the EU. 

While autonomous vehicle technology appears to be developing at a fast pace, no commercially available vehicles have yet reached the stability and reliability required for the highest level of driving autonomy~\cite{saeAutonomi}. An autonomous car, to safely operate in an unknown environment, requires multiple components to be correctly tuned and tested. Obstacles detection and sensor fusion algorithms need to be reliable in numerous scenarios, and redundancy of the sensors should guarantee correct detection from at least one sensor while the vehicle is driving. Moreover, state estimators and planning algorithms have to be tested in multiple conditions, like scenarios where GPS data might not be available or provide false detection and cameras might be blinded by low sunlight. All those scenarios can be easily tested in a simulated environment. In the last years, we indeed witnessed the rise in popularity of accurate autonomous driving simulators, like Carla~\cite{Dosovitskiy17}. Nevertheless, it is also important to validate each solution in real scenarios since simulated environments will never fully replicate all the aspects of the real world. It is challenging to simulate sensors' noise, changes in the environment, and other agents' behavior. 

For those reasons, while software industries developed accurate simulators, research groups also focused on building development platforms for autonomous vehicles. While the first working prototypes date back to the eighties~\cite{dickmanns1987autonomous}, the first significant improvement came with the DARPA Grand Challenge. This competition incentivized the development of real autonomous vehicles from multiple universities and research groups~\cite{thrun2006stanley},~\cite{urmson2008autonomous}. Later other competitions kept the interest on this topic alive, from the VisLab Intercontinental Autonomous Challenge~\cite{broggi2012vislab} to the PROUD—Public Road Urban Driverless-Car Test~\cite{broggi2015proud}.
More recently, we witnessed an increased interest in the topics of autonomous vehicles. Many companies started developing their setup~\cite{yurtsever2020survey}, which in most scenarios consists of a multi-sensors rig with cameras, lidars, and radars. This has been the standard configuration on which newer prototypes from smaller research group has also settled in~\cite{ferranti2019safevru}. 

\begin{figure}[t!]
	\centering
	{\includegraphics[width=0.48\textwidth]{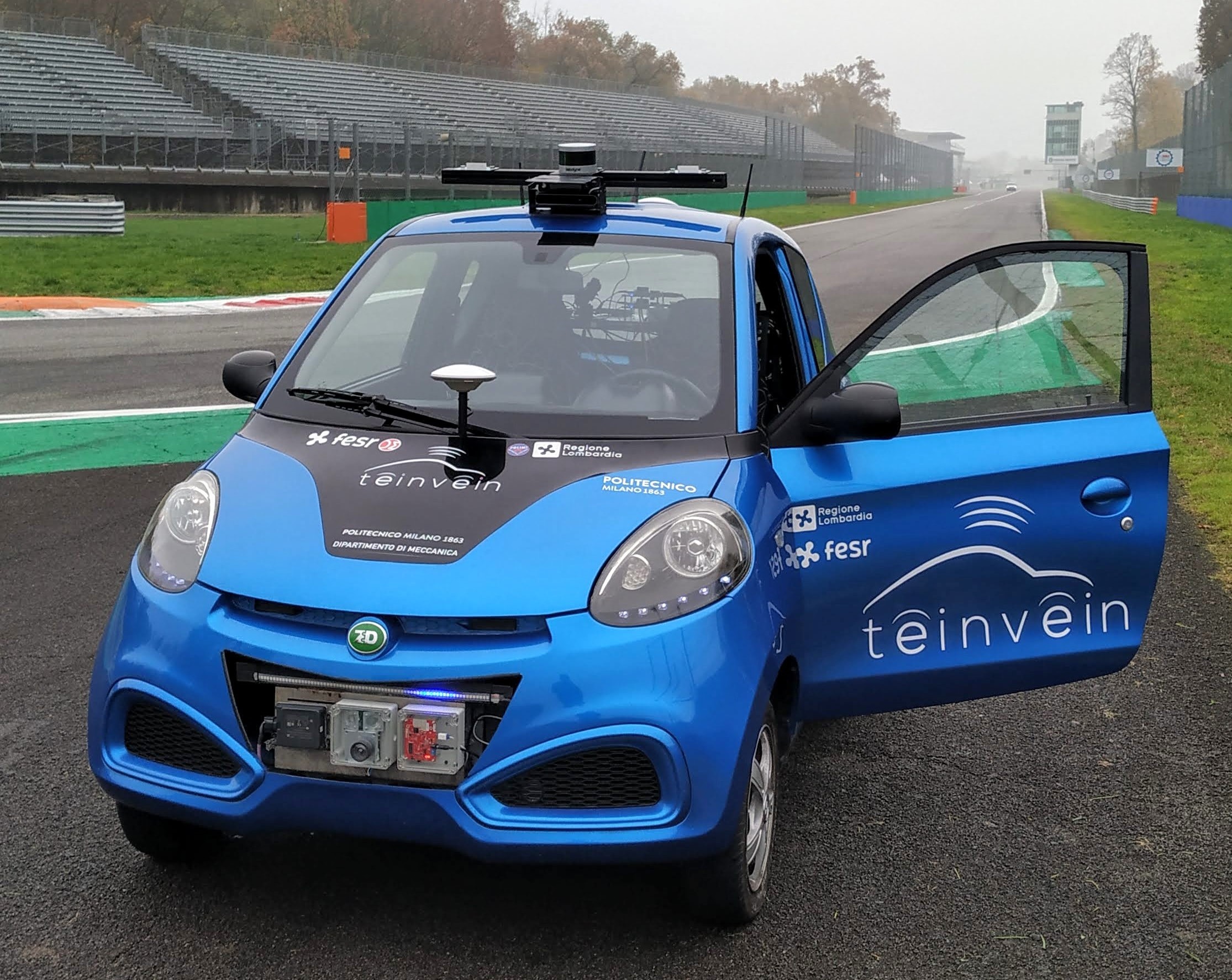}}
	\caption{Image of the prototypical vehicle.}
	\label{fig:car}
\end{figure}

What we are proposing in this paper is a prototypical platform, Fig.~\ref{fig:car}, based on a commercial vehicle, for autonomous vehicles. Our solution can be employed to test AV control and planning algorithms, obstacle detection, tracking solutions, and Infrastructure to Vehicle (I2V) communication. Moreover, thanks to the complete sensor suite, it is a perfect platform for datasets acquisition, one of the core components of each testing pipeline. Finally, since the vehicle can perform autonomously challenging tasks in a complex environment and thanks to modular and expandable hardware architecture, it is well suited for other interdisciplinary studies, more aimed at the final autonomous vehicle product, like analyzing passenger's stress and acceptance by using physiologic measurements.

Section \ref{sec:VA} will describe the overall architecture of the vehicle in terms of tasks, functionalities, and logical interactions between the different components. In Section \ref{sec:HW} all mechanical intervention for sensing, actuation, and computational purposes is presented, while in section \ref{sec:SW} the main features of the different algorithms that make up the overall autonomous driving control system are described. In section \ref{sec:tests} a testing scenario is presented, and experimental tests are reported and commented. Finally, in section~\ref{sec:CONC} the conclusions are drawn.

\section{Vehicle architecture}\label{sec:VA}
% System architecture?

The vehicle adopted, as an initial mechanical platform for our prototype, is a fully electric micro-car provided by Share'NGo\textsuperscript{\textregistered}, a  light quadricycle used for car-sharing mobility services.

\begin{figure}[t!]
	\centering
	{\includegraphics[width=0.45\textwidth]{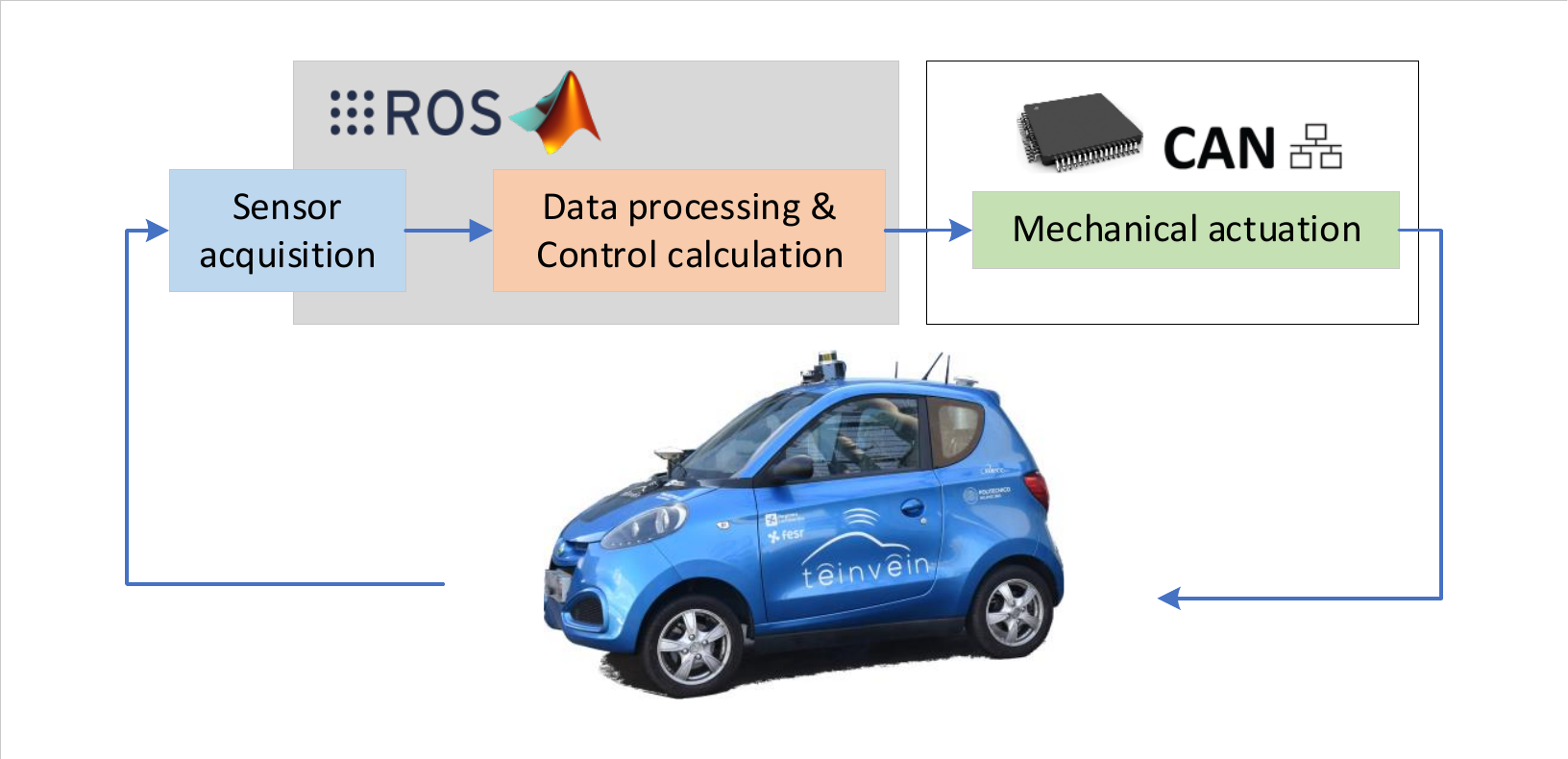}}
	\caption{Schematic of logical components of vehicle architecture.}
	\label{fig:veh_architecture}
\end{figure}
As shown in fig. \ref{fig:veh_architecture}, the overall architecture of the prototype can be split into three main tasks: raw data acquisition from sensor set-up mounted on the vehicle; data processing and control strategy algorithms and finally the mechanical actuation of the vehicle.

The software architecture adopted is distributed and can be mainly divided into two parts:
\begin{itemize}
    \item \textbf{a soft real-time part}: it is responsible for data acquisition from all physical sensors, data processing (image processing, state estimation, and obstacle tracking), and motion control. it is distributed on different computing units, and it is implemented on ROS middleware framework~\cite{quigley2009ros}; 
    \item \textbf{a hard real-time part}: it is responsible for ego vehicle data acquisition as vehicle speed and steering angle as well as of computing control commands for the actuators. it is implemented in an embedded microcontroller board.  
\end{itemize} 

\section{Hardware setup}\label{sec:HW}

The main mechanical interventions performed on the commercial vehicle can be classified as actuators remoting and sensors setup.
Actuators remoting involves intervention on the steering system and the powertrain to remotely control the vehicle.
The prototype car is already equipped with an electric power-steering DC motor. In order to properly control the steering angle, the DC motor was controlled through a control algorithm that acts on the angular position of the steering system through a feedback control action based on measurements of an incremental encoder added to the assembly to measure the steering angle.

On the powertrain side, the vehicle's gas pedal consists of an active system that generates a digital signal to the internal ECU of the vehicle. An electrical switch has been mounted between the ECU and the gas pedal. When required, a control board can produce a digital signal that replicates the one usually coming from the original gas pedal.

\begin{figure}[t!]
	\centering
	{\includegraphics[width=0.4\textwidth]{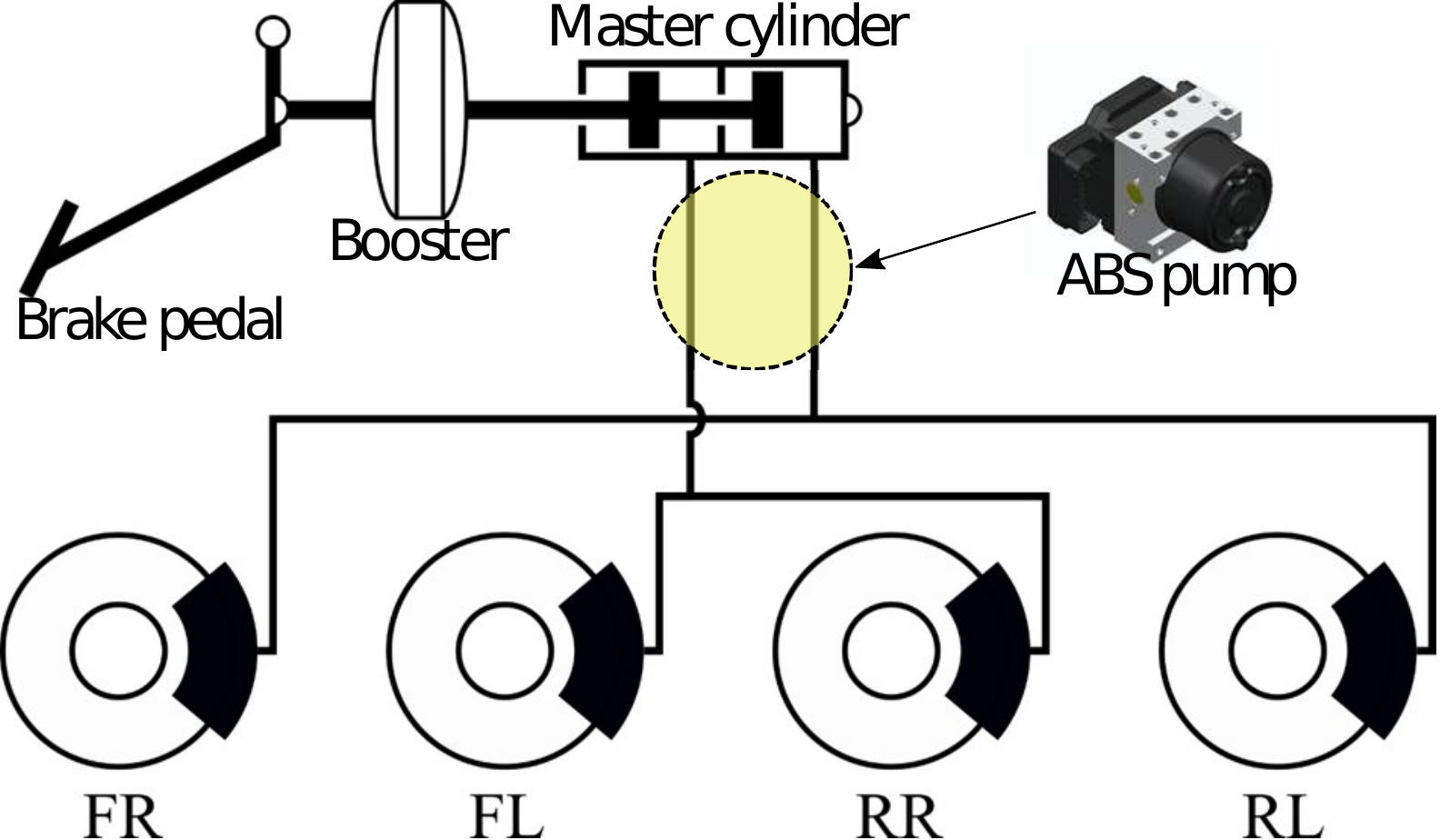}}
	\caption[Schematic of the modified braking system.]{Schematic of the modified braking system.}
	\label{fig:braking_sys}
\end{figure}

 Finally, in order to properly brake, an ABS control board is included in the original baking system between the braking system master cylinder and the four brake calipers, as shown in fig.\ref{fig:braking_sys}. The ABS ECU included was modified and can be directly controlled to modulate its output pressure by acting on the voltage applied to the hydraulic pump and the solenoid valve.

Actuators' input commands are provided as the result of feedback control algorithms. Feedback control loops are designed in order to minimize the error between reference steering angle and longitudinal speed (input values provided by motion planning algorithms) and actual steering angle and longitudinal speed. Where actual vehicle speed is estimated by means of two encoders mounted on the wheel hubs. 
As already introduced, the actuators control board, which consists of a micro-controller unit, is also responsible for communicating actual vehicle speed and steering angle to the other soft real-time boards.

%\section{Sensors}\label{sec:Sen}

\begin{figure}[t!]
	\centering
	{\includegraphics[width=0.5\textwidth]{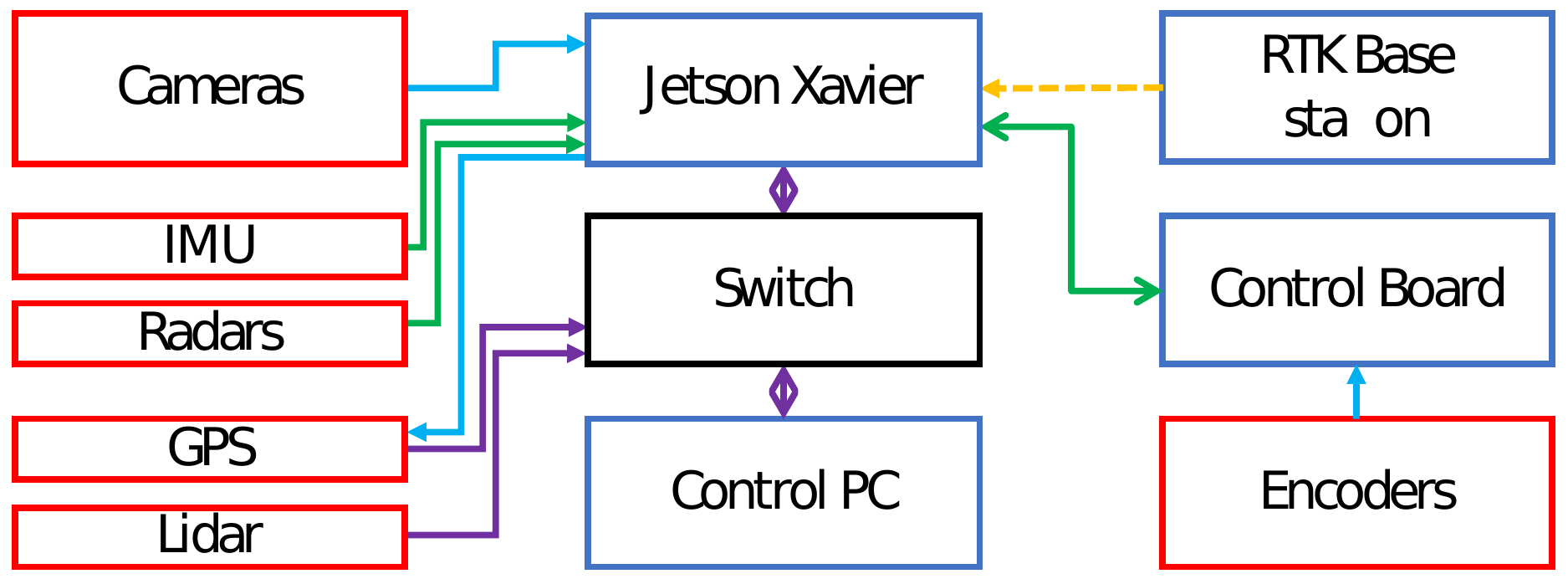}}
	\caption{Schematic of the system connections. Red boxes indicates sensors, blue ones computation unit. Purple connection represents Ethernet, yellow 4G, light blue serial and USB, and green CAN.}
	\label{fig:sensors}
\end{figure}

%Sensors setup is described in the following.
As shown in Fig.~\ref{fig:car} the vehicle is equipped with multiple sensors, each one with specific features, interfaces, and running frequencies. For this reason, it is not possible to acquire and process all data using only one device and a common communication interface. Still, multiple solutions have been adopted, as shown in Fig.~\ref{fig:sensors}.

In particular, sensors providing information on the vehicle state works at high frequency but with a small bandwidth. Therefore are connected to the main processing unit, a Jetson Xavier, using CAN and ethernet. Of particular interest is the communication between the GPS sensors on the vehicle and the RTK correction base station. Usually, the correction messages from the base station are transmitted using radio waves. This approach is feasible only if the maximum distance is a few hundred meters due to the limited power of the commercially available transmitters. Since the experimental vehicle has to operate on a larger space, the correction messages are transmitted using a 4G connection through a VPN as standard ROS messages. In such a way, the prototype's computing platform can subscribe to the base station's messages and forward them to the GPS sensor.
Ethernet connection is also used to acquire the lidar data. Since the vehicle is equipped with only one Lidar, this solution is viable, and the data rate through the switch is never too high. Finally, the communication between the processing unit and the control pc, which runs the Matlab logics, is also performed using an ethernet connection. Contrary data from cameras are too heavy to be transmitted through ethernet. For this reason, all cameras are connected through USB-3.0 to the Jetson PC. Radar data also have limited bandwidth since the employed sensors are only 2D. Therefore it is possible to acquire them using the CAN network without any risk of saturating it. 

Going further into details, the experimental vehicle is equipped with two Swiftnav-RTK GPS receivers, one mounted on the front of the car and one on the back. This configuration allows an accurate estimate of the vehicle state and possibly a differential setup to compute the prototype heading. The GPS board is already equipped with IMU and magnetometer. An additional automotive IMU (Bosh MM5.10) is mounted in correspondence with the vehicle's center of gravity to provide the state estimator with better data. 
The core of the obstacle detection system is a Velodyne VLP-16 Lidar mounted on the roof, slightly tilted in the forward direction. A couple of Continental ARS 401 Radar are mounted in the front and rear bumper of the vehicle to get the crucial information of obstacles speed. Finally, a ZED Stereolabs camera is mounted near facing the moving direction, while multiple fisheye cameras are integrated into the vehicle to guarantee a 360\degree coverage of the vehicle surrounding.

\section{Software architecture}\label{sec:SW}

The distributed computing system shown in Fig.~\ref{fig:veh_architecture} and Fig.~\ref{fig:sensors}, consists of a GPU-based computing board (Nvidia Jetson Xavier) and a commercial x86 laptop. It represents the ``brain" of the vehicle and it's responsible for a reliable surrounding awareness and it has to evaluate and decide driving tasks properly. 
%% obstalces and ego-localization
In order to navigate in an unknown environment, the autonomous vehicle needs a detailed representation of the surrounding. In particular, planning algorithms require a list of fully characterized obstacles (i.e., position, size, speed, class). Since the only sensor providing a 360\degree is the Lidar, all other data are aligned to it. 
In order to do that, on the Jetson Xavier board, the pointcloud coming from the Velodyne is processed to remove the ground plane and projected to a 2D grid to retrieve an occupancy grid, on which it is possible to extract obstacles clusters with limited computational power requirements~\cite{8804556}. Data from other sensors are then fused to provide a more accurate representation of each obstacle. In particular, data from the radar is used to estimate the obstacle position better and provide crucial information on speed. Instead, camera images are processed with a convolutional neural network~\cite{7780460} to retrieve 2D bounding boxes on the image plane. Those boxes are then reprojected through bird-eye-view on the ground plane to assign to each obstacle a class. Finally, the state of each obstacle, described as position, orientation, and speed, is tracked using an Unscented Kalman Filter (UKF) to account for the strong nonlinearities of the system. Ego-state estimation is fundamental information for motion planning algorithms. This task is implemented using MATLAB programming language on the x86 laptop. Several techniques have been evaluated by the authors as reported in ~\cite{8804527}, and finally, a UKF algorithm was chosen. In particular, the solution adopted makes use of the road reference frame both for obstacle tracking and ego vehicle state estimation instead of a more traditional Cartesian reference frame on which sensors provide data. This allows an accurate estimate of the position of each obstacle and ego vehicle in the same reference frame used by the planning algorithm, removing the need for an intermediate conversion which would introduce approximation errors and delays~\cite{BERSANI2021103662}. 
%% logica di controllo
A common approach in literature to deal with the overall driving problem is to decouple it into a sequence of different tasks by means of a hierarchical structure composed of different layers \cite{7490340,6839646} as reported in fig. [schema]. A motion planner based on a Nonlinear Model Predictive Control (NMPC) algorithm was developed according to that approach. Route planning and Behavioral tasks are instead neglected by reducing the set of scenarios that can be managed. In particular, since the aim of the work proposed in this paper is to show and to prove the performances of the prototype vehicle, a simple trajectory following scenario is considered where the reference path is pre-calculated, and the vehicle is expected to drive on the lane close to its cruising speed. The motion planner algorithm is capable of dealing with road boundaries and obstacles (both static and moving) thanks to soft and hard constraints and to follow the centerline of the road lane minimizing lateral error and the deviation from cruising speed. It used information provided by ego-localization and obstacle tracking algorithms as inputs to calculate an optimal and safety trajectory as output.
%% descrizione più ciccia del controllore? magari qualche formula??
More into detail, the NMPC formulation adopted solves an Optmization Constrained Problem at each time iteration, in the general form reported in \eqref{eq:typicalOCP}:
\begin{equation}
        \label{eq:typicalOCP}
        \begin{aligned}
        \underset{x(.),u(.)}{minimize} \int_{0}^{T}L(x(t),u(t))dt \; + \; &E(x(T))\\    
        subject\:to\hspace{5mm} \dot{x}(t) - f(x(t),u(t)) &= 0, \\
        h(x(t),u(t)) &\leq 0, \\
        r(x(T)) &\leq 0 \; 
        \end{aligned}
\end{equation}
where \emph{\textbf{L}} is the objective function that considers deviation from reference trajectory, cruising speed, relative distance with detected obstacles, presence of speed bumps as well as control input penalty, while \emph{\textbf{E}} is the terminal cost function. Finally the optimization problem is constrained by vehicle model equation $\dot{x}(t) = f(x(t),u(t))$ and physical constraint function\emph{\textbf{h}}, which represents state limits (geometrical and physical limits) as well as control action limits. In \cite{8493215,micheli2021nmpc} the starting point for the motion planner mathematical formulation integrated on the vehicle is reported.

Starting from the results provided by the motion planner, at each time step a set of optimal reference values of steering angle and longitudinal speed $\delta_r,V_r$ is provided. These values are converted into a CAN message and communicated to the actuators' control board.

\section{experimental tests}\label{sec:tests}
\begin{figure}[t!]
	\centering
	{\includegraphics[width=0.48\textwidth]{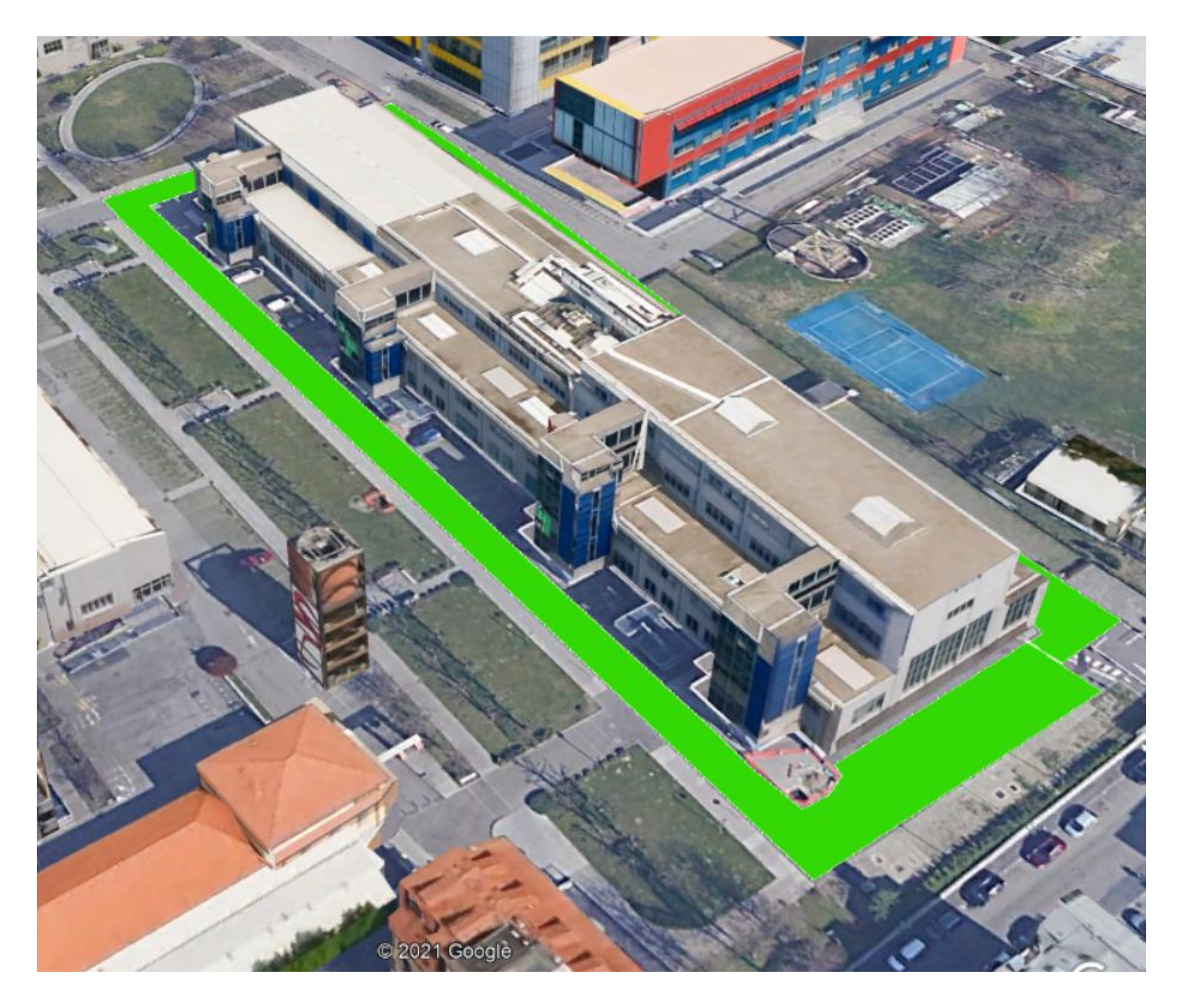}}
	\caption{3D view of the test track. Green area represents the driving lane}
	\label{fig:areatest}
\end{figure}

The prototype vehicle has been tested in several scenarios. In this paper, an experimental test is reported and deeply analyzed in order to prove and evaluate the performances of the prototype platform proposed. In particular, to comply with insurance and safety rules, a single road lane surrounding a building inside the Politecnico di Milano campus is chosen as a testing stage; the area is shown in Fig.~\ref{fig:areatest}. This area, highlighted in green, represents a hypothetical single lane used as a reference path for the motion planner algorithm.

The total road width varies between $6\,m$ and $6.6\,m$, but for safety reasons and to consider the vehicle's width, the maximum lateral displacement of CoG is numerically limited to $4.6\,m$ inside the algorithm. The overall length of the track is equal to  $364\,m$ and it is composed of two long straight-lines of around  $135\,m$ and two short straight-lines of $45\,m$ linked by four sharp curves. Moreover, an obstacle placed in the middle of the second straight-line is considered during the test (only during the first lap). This scenario mimics a typical urban area characterized by sharp turns and surrounded by high buildings and threes that can limit the accuracy and reliability of GPS signals.

The experimental test conducted aims at providing a smooth driving within the track limits and a longitudinal speed close to the cruising speed ($3\,m/s$) on straight-line and with a limited lateral acceleration as prescribed by the formulation of OCP. The data reported cover two complete laps. 

\begin{figure}[t!]
	\centering
	{\includegraphics[width=0.45\textwidth]{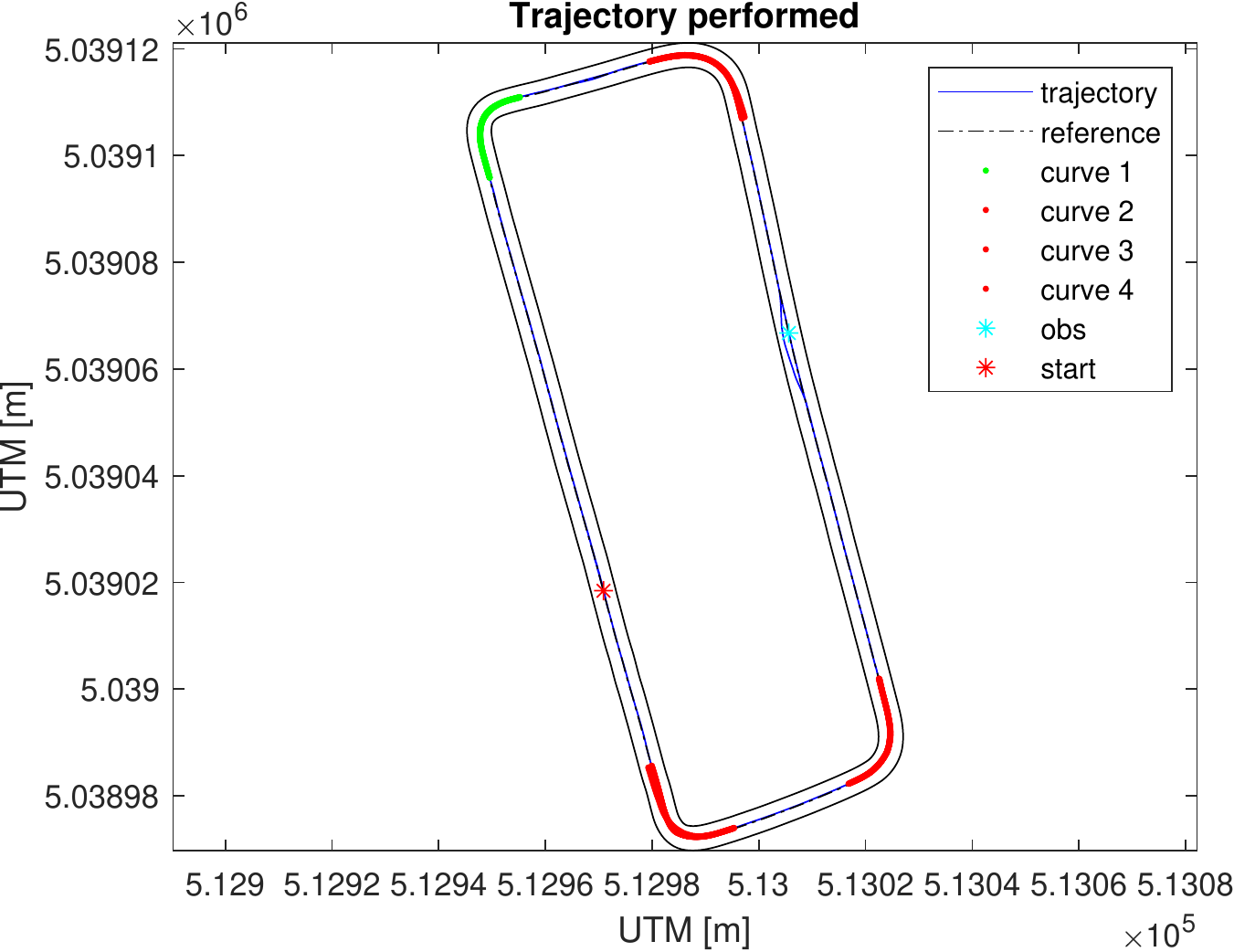}}
	\caption{trajectory of the vehicle over four laps}
	\label{fig:traiettorie}
\end{figure}
In Fig.~\ref{fig:traiettorie} the vehicle position on the track during experimental tests is reported as the blue dotted line, while the black line represents the ideal centerline and lateral bounds considered by the controller. The green line represents the first turn, while the red lines are the other ones. Finally, the red dot represents the starting line of the track, and the cyan dot is the position of the obstacle (a circular object with a radius of $0.7\,m$ considered only in the first lap performed). As visible, the vehicle is capable of performing a proper driving trajectory well within the road bounds. 
\begin{figure}[t!]
	\centering
	{\includegraphics[width=0.5\textwidth]{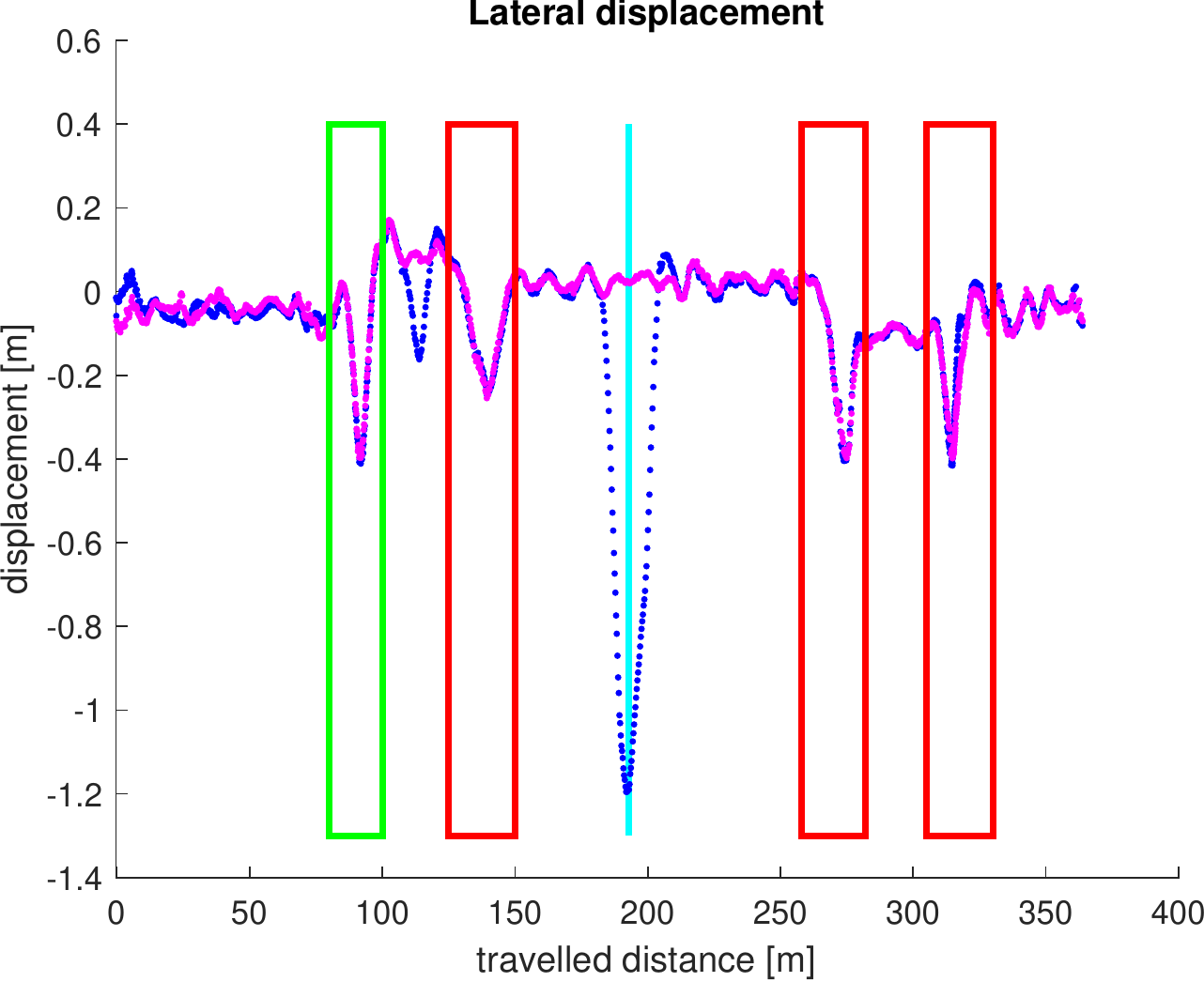}}
	\caption{lateral displacement over four laps over travelled distance}
	\label{fig:latdisp}
\end{figure}
In detail, Fig.~\ref{fig:latdisp} shows the lateral deviation with respect to the centerline for the two laps performed as blue and magenta lines, respectively. The results show a maximum displacement of around $0.4\,m$ through the corners while its RMS value is around $0.065\,m$ considering straight lines (without obstacle). It is important to notice how those values are always within the road boundaries imposed by the controller. Therefore this deviation from the exact centerline has not to be considered as an error but as a choice of the control algorithm to perform smoother turns. In particular, the highest offset can be registered in the tight corners, where the turning space is limited. For this reason, the vehicle moves away from the exact centerline, like a human driver would do, to guarantee a more comfortable and less sharp right turn. The only significative deviation from the centerline can be identified after around $200\, m$ when the vehicle has to avoid an obstacle. As shown in the figure in correspondence of the cyan line in the first lap, represented in blue, the car had to move significantly away from the ideal trajectory, with a lateral displacement of $ 1.2\, m$ to avoid the obstacle. Nevertheless, the control algorithm handled the scenario and moved the vehicle back to the correct centerline after the maneuver.
\begin{figure}[t!]
	\centering
	{\includegraphics[width=0.5\textwidth]{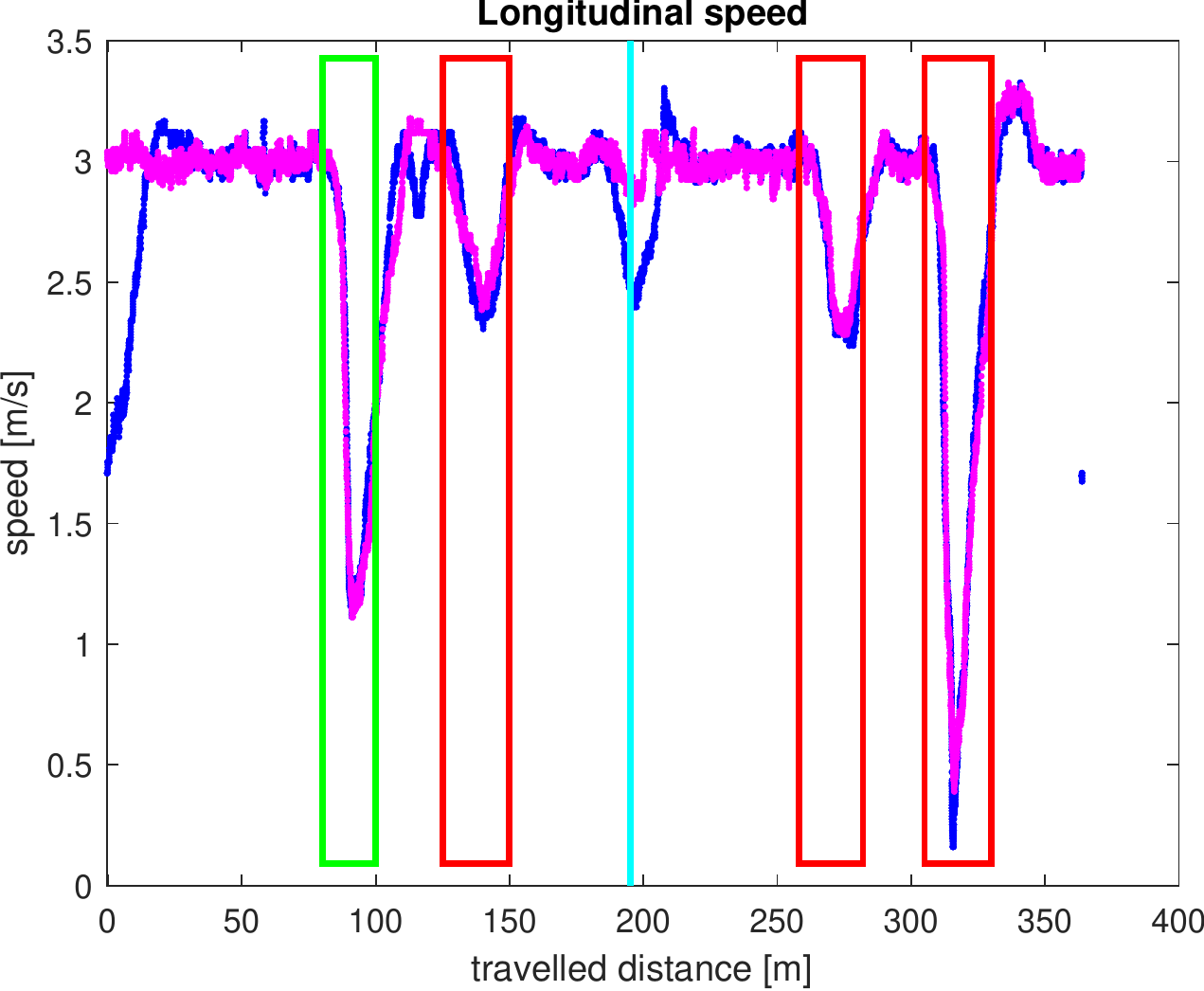}}
	\caption{longitudinal speed over four laps over travelled distance}
	\label{fig:longsp}
\end{figure}
The longitudinal speed profile is reported in Fig.~\ref{fig:longsp}. As visible, the speed along the straight lines coincides with the cruising speed of $3\, m/s$, with minimal fluctuations. In particular, Cruising error RMS along straight lines (excluding starting straight line where vehicle start from standstill and straight line with the obstacle) is equal to $0.09\,m/s$. Instead, the motion planner reduces the longitudinal speed during curve phases to limit the lateral acceleration and provide a more comfortable ride to the passenger. In this scenario, it is also possible to notice how the speed significantly decreases when the experimental vehicle approaches the obstacle at $200\, m $ on the straight. Similar to what has been shown for the lateral displacement, the control algorithm returns to the original configuration and accelerates back to the correct cruising speed of $3\,m/s$ after surpassing the obstacle.

\begin{figure}[t!]
	\centering
	{\includegraphics[width=0.5\textwidth]{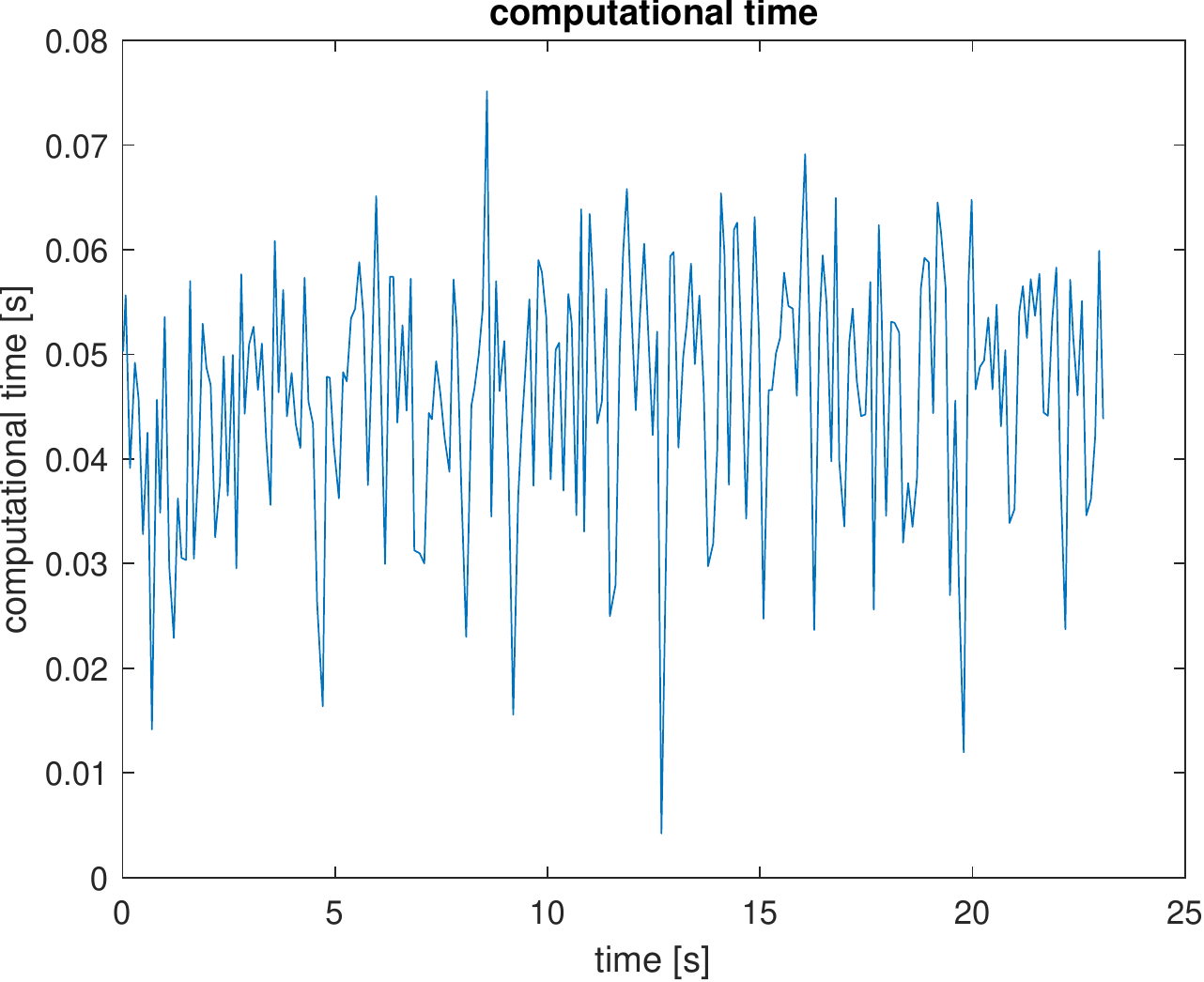}}
	\caption{computational time of state estimator + motion planner}
	\label{fig:comptime}
\end{figure}

%% non ho dati di questo
In Fig.~\ref{fig:comptime} combined computational time of state estimator and motion planner algorithms is reported. As visible by the graph, the computational time for the test reported always remains below $0.075\,s$ which means that it can properly behave when the working frequency is set to $10\,Hz$

\begin{figure}[t!]
	\centering
	{\includegraphics[width=0.5\textwidth]{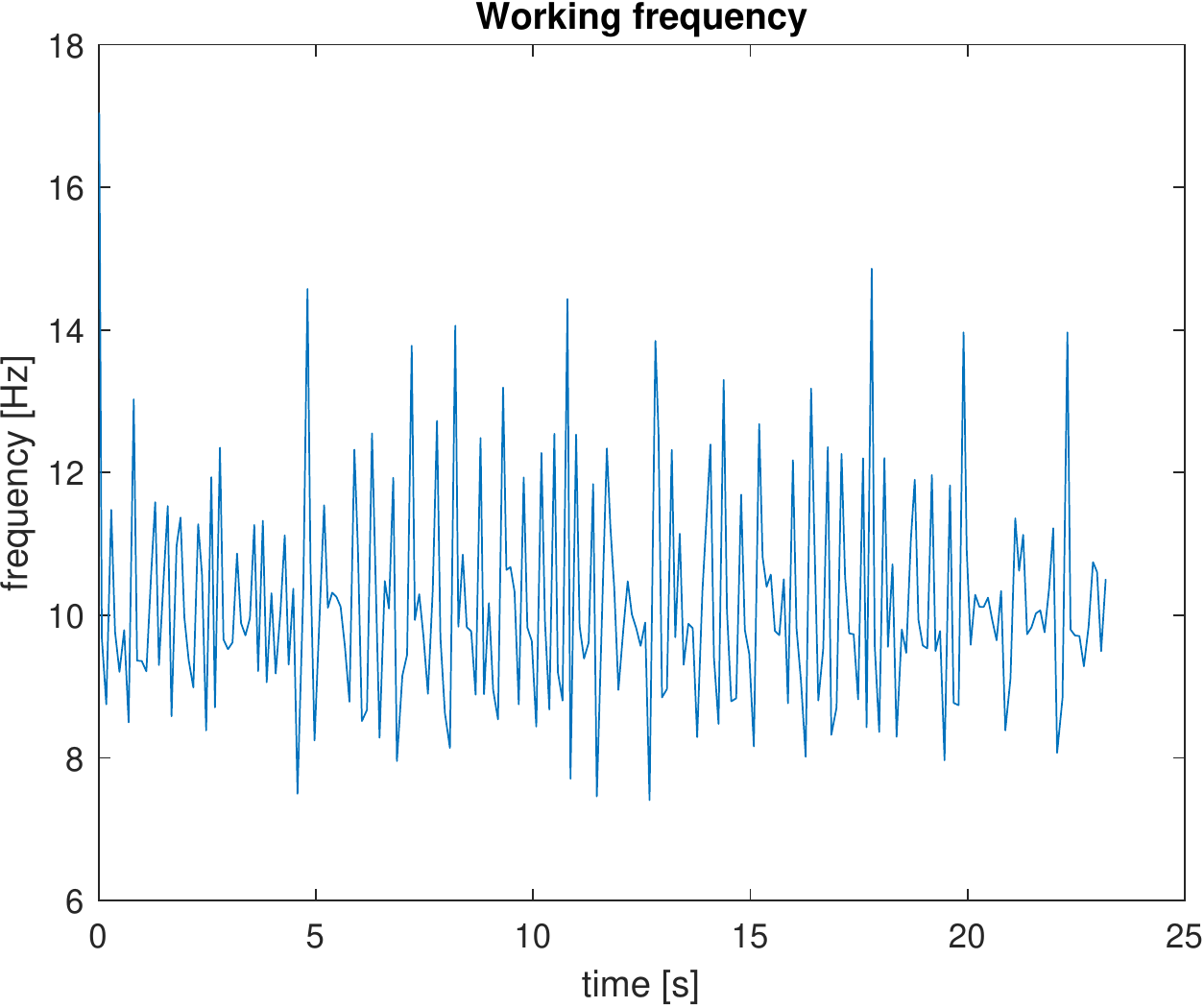}}
	\caption{working frequency of the algorithm during  experimental tests}
	\label{fig:workfr}
\end{figure}

Finally, in Fig.~\ref{fig:workfr} the frequency of the control messages provided by soft real-time part is shown.
This value considers the overall computation time of all algorithms, data acquisition, and the "waiting time" to correctly compute an output with a fixed frequency. As visible in the figure, working frequency presents a mean value around  $10\,Hz$ as expected. It is also visible a small fluctuation due to the software architecture adopted, based on ROS, which cannot guarantee a stable frequency. This is not a problem since the hard real-time part of the architecture can deal with this phenomenon. The mean value of the working frequency is the one imposed on the system: this proves that the overall computation time of all algorithms and data acquisition is less than or at least equal to $0.1\,s$ as required for a real-time implementation of the scenario under analysis.

\section{Conclusion}\label{sec:CONC}
In this paper, our prototype platform for experimental research on connected and autonomous driving projects is presented. In detail, the overall architecture comprehensive of physical aspects like actuators remoting and sensors set-up as well as a comprehensive description of the main algorithms required for ego-localization, environment perception, motion planning, and actuation is fully described. Experimental results obtained in a real-world test conducted in an urban-like scenario show satisfying performances of the prototype with a limited lateral deviation always between road bounds, capable of properly maintain a cruising speed imposed and real-time capabilities of the overall architecture at a tested control frequency of $10\,Hz$.
The research platform presented is a fundamental test bench for ongoing projects related to control design for autonomous and connected vehicles and passenger's stress and acceptance evaluation.

\section*{Acknowledgment}
The Italian Ministry of Education, University and Research is acknowledged for the support provided through the Project "Department of Excellence LIS4.0 - Lightweight and Smart Structures for Industry 4.0”.

\bibliographystyle{IEEEtran}
\bibliography{main.bbl}

%\bibliography{Arrigoni_bib_AV} % IEEEabrv,BiblioBUS,
\end{document}